\documentclass[conference]{IEEEtran}
\IEEEoverridecommandlockouts
\usepackage{cite}
\usepackage{amsmath,amssymb,amsfonts}
\usepackage{algorithmic}
\usepackage{graphicx}
\usepackage{textcomp}
\usepackage{hyperref}
\usepackage{float}
\usepackage{booktabs}
\usepackage{multirow}
\usepackage{caption}
\usepackage{subfigure}
\usepackage{tcolorbox}
\tcbuselibrary{listingsutf8}
\newtcolorbox[auto counter]{promptbox}[2][]{%
  colback=gray!5!white, 
  colframe=gray!75!black, 
  fonttitle=\bfseries,
  fontupper=\footnotesize,
  title=Prompt~\thetcbcounter: #2, 
  label={#1}
}
\def\BibTeX{{\rm B\kern-.05em{\sc i\kern-.025em b}\kern-.08em
    T\kern-.1667em\lower.7ex\hbox{E}\kern-.125emX}}
\begin{document}

\title{Tabular Data with Class Imbalance: Predicting Electric Vehicle Crash Severity with Pretrained Transformers (TabPFN) and Mamba-Based Models}

\author{\IEEEauthorblockN{
Shriyank Somvanshi\IEEEauthorrefmark{1}, 
Pavan Hebli\IEEEauthorrefmark{1}, 
Gaurab Chhetri\IEEEauthorrefmark{2}, 
Subasish Das, Ph.D.\IEEEauthorrefmark{1}
}
\IEEEauthorblockA{\IEEEauthorrefmark{1}Ingram School of Engineering, Texas State University, San Marcos, Texas, USA\\
Email: \{shriyank, zea16, subasish\}@txstate.edu}
\IEEEauthorblockA{\IEEEauthorrefmark{2}College of Science and Engineering, Texas State University, San Marcos, Texas, USA\\
Email: gaurab@txstate.edu}
}

\maketitle

\begin{abstract}
This study presents a deep tabular learning framework for predicting crash severity in electric vehicle (EV) collisions using real-world crash data from Texas (2017-2023). After filtering for electric-only vehicles, 23,301 EV-involved crash records were analyzed. Feature importance techniques using XGBoost and Random Forest identified intersection relation, first harmful event, person age, crash speed limit, and day of week as the top predictors, along with advanced safety features like automatic emergency braking. To address class imbalance, Synthetic Minority Over-sampling Technique and Edited Nearest Neighbors (SMOTEENN) resampling was applied. Three state-of-the-art deep tabular models, TabPFN, MambaNet, and MambaAttention, were benchmarked for severity prediction. While TabPFN demonstrated strong generalization, MambaAttention achieved superior performance in classifying severe injury cases due to its attention-based feature reweighting. The findings highlight the potential of deep tabular architectures for improving crash severity prediction and enabling data-driven safety interventions in EV crash contexts.

\end{abstract}


\section{Introduction}

\subsection{Motivation and Background}
The rapid adoption of electric vehicles (EVs) introduces unique safety challenges due to differences in vehicle dynamics, quiet operation, and battery-related risks compared to traditional vehicles \cite{moore2020sound, yi2021functional, in2025development}. Existing safety standards and predictive models often treat EVs and conventional vehicles similarly, despite the distinct crash behaviors of EVs \cite{aziz2024road}. Accurately predicting crash severity for EVs is crucial for developing proactive safety interventions and real-time decision support systems.

\subsection{Research Gap}
Most existing crash severity studies rely on statistical models or traditional machine learning methods \cite{yang2022predicting}, which are limited in capturing the complex, non-linear interactions present in EV crash data. While ensemble models like XGBoost have shown promise \cite{yang2022predicting}, there is limited research applying advanced deep tabular learning models, such as TabPFN, MambaNet, or attention-based architectures—to EV-specific crash datasets \cite{rafe2024comparative, somvanshi2025applying}. Furthermore, modern safety features such as Automatic Emergency Braking (AEB) and driver assistance systems are rarely incorporated into severity prediction models, leaving a gap in leveraging the latest vehicle technology data \cite{sequeira2021investigation}.

\subsection{Study Objective}
This study aims to develop a deep tabular learning framework (see Figure 1) for predicting EV crash severity using structured crash data enriched with advanced safety feature information. We evaluate models including TabPFN, MambaNet, and MambaAttention, alongside feature selection using ensemble-based methods (XGBoost and Random Forest), and apply SMOTEENN to address class imbalance. The objective is to benchmark these models and identify key predictors for EV crash outcomes.

\subsection{Contribution Summary}
This study proposes a specialized framework for predicting EV crash severity using advanced deep tabular models, including TabPFN and Mamba-based architectures. A hybrid feature selection method combining Random Forest and XGBoost identifies key predictors, while SMOTEENN addresses class imbalance. Comprehensive evaluation metrics and interpretability analysis highlight the effectiveness of attention-based models in improving prediction accuracy and informing proactive safety measures.

 \begin{figure*}[h!]
\centering
\includegraphics[width=1\linewidth]{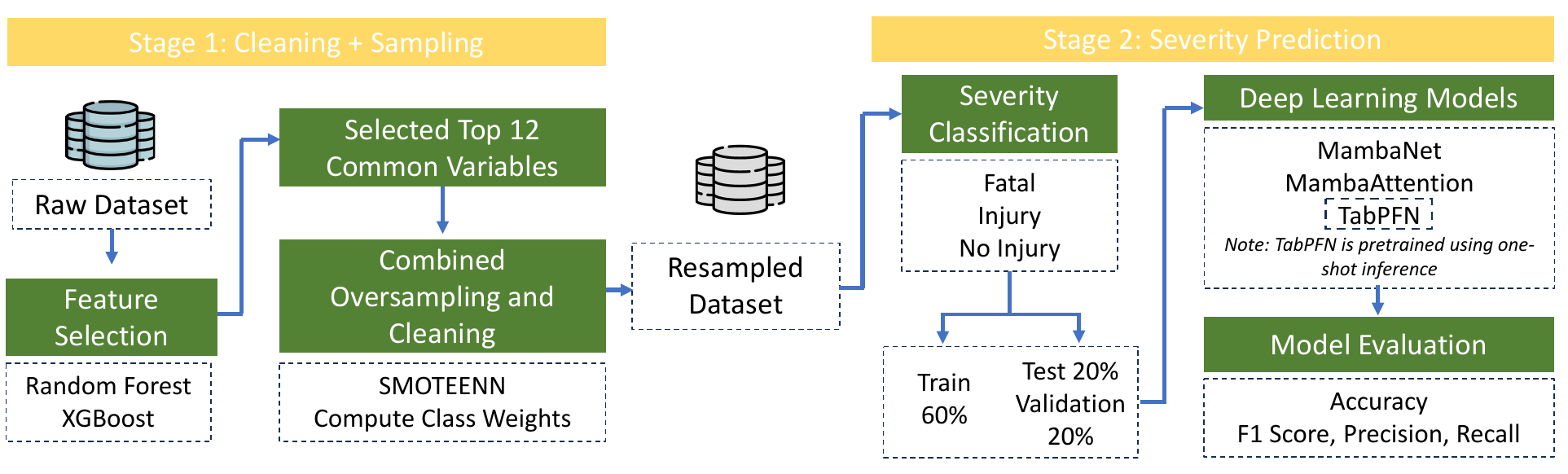}
\caption{Study Design}
\label{fig:flowchat}
\end{figure*}

\section{Literature Review}

\subsection{Crash Severity Prediction with Machine Learning}

Traditional crash severity studies relied on statistical models such as logistic regression and ordered logit models, which struggle to capture non-linear relationships in high-dimensional crash data \cite{DAS2018114, das2022artificial}. Machine learning approaches—such as decision trees, random forests, and XGBoost, have demonstrated superior predictive performance by effectively modeling these complex patterns \cite{chakraborty2023causal}. Among these, ensemble methods like XGBoost have consistently outperformed conventional models in terms of accuracy and feature interpretability, highlighting key predictors such as collision type, weather, and road surface conditions \cite{scarano2025cyclist}.

\subsection{Deep Learning for Tabular Data}

Recent advances in deep learning have produced competitive alternatives to gradient-boosted trees for structured data \cite{somvanshi2024a, a32}. Notably, TabPFN, a pretrained transformer tailored for small-scale tabular classification, offers state-of-the-art performance with no hyperparameter tuning via one-shot inference and in-context learning that approximates Bayesian inference \cite{hollmann2022tabpfn}. Transformer-based models such as FT-Transformer and SAINT further enhance tabular learning by employing self-attention and intersample attention to model feature- and sample-level dependencies, with SAINT achieving performance on par with gradient boosting by effectively embedding both categorical and continuous variables \cite{gorishniy2021revisiting, somepalli2021saint}.

\subsection{Mamba-based Architectures}

State Space Models (SSMs) have recently been adapted for tabular data. MambaTab, derived from the Mamba architecture, treats tabular inputs as sequences and leverages linear-time sequence modeling, offering strong performance with fewer parameters \cite{thielmann2024mambular, ahamed2024mambatab, somvanshi2025from}. Hybrid architectures like MambaNet integrate CNNs and state-space layers to capture feature dependencies, while attention-enhanced variants such as MambaAttention have shown superior classification performance by dynamically reweighting key features.

\subsection{Class Imbalance Handling}

Crash datasets often suffer from severe class imbalance, where non-injury cases dominate. To mitigate this, SMOTEENN, a hybrid of Synthetic Minority Oversampling (SMOTE) and Edited Nearest Neighbors (ENN), has proven effective in balancing data while removing noise, improving model performance across metrics such as accuracy and F1-score \cite{husain2025smote}. This approach enhances deep models’ ability to correctly classify minority classes like severe and fatal crashes.

\subsection{Gap in Literature: EV-Specific Crash Modeling}

Although EVs are increasingly adopted, studies explicitly focused on EV-specific crash severity remain scarce. Most research either aggregates EVs with conventional vehicles or centers on internal combustion engine crashes, overlooking the distinct crash dynamics of EVs such as silent operation, high torque, and battery weight. A recent study using Virginia crash data found EVs to be associated with a 14\% reduction in severe injury odds, but few analyses isolate EV crashes for detailed modeling \cite{zhaielectric}. Other work, such as in Metro Manila, applied machine learning and spatio-temporal tools to show that EV crashes are more frequent in high-density, off-peak settings lacking EV infrastructure \cite{salang2024spatio}. Additionally, the role of advanced safety systems, particularly automatic emergency braking (AEB), remains underexplored in EV contexts, despite evidence of their growing real-world effectiveness from 46\% to 52\% between 2015–2023 \cite{cicchino2017effectiveness, aukema2025study}.

\section{Data Preparation}

\subsection{Dataset Description}

This study utilized crash data from the Texas Crash Records Information System (CRIS), spanning the years 2017 to 2023. CRIS is a statewide database maintained by the Texas Department of Transportation (TxDOT) for all reportable motor vehicle traffic crashes, as mandated by Texas Transportation Code Chapter 550 \cite{txdot_crash_records}. The dataset was filtered to include only electric vehicle (EV) crashes by selecting records where the \textit{IsElectric} flag was \textit{True}, resulting in 23,301 EV-involved crash records.

The target variable for this study is \textit{Prsn\_Injry\_Sev\_ID}, representing injury severity. Based on the KABCO scale, Fatal (K), Suspected Serious (A), Minor (B), Possible (C), and No Injury (O) \cite{txdot_crash_records}, we reclassified outcomes into three categories: Fatal/Severe (KA), Moderate/Minor (BC), and No Injury (O). This grouping aligns with national reporting standards and enhances model interpretability.

The dataset comprises 34 variables covering demographics (\textit{Prsn\_Age}, \textit{Prsn\_Gndr\_ID}), vehicle details (\textit{Veh\_Make\_ID}, \textit{Veh\_Body\_Styl\_ID}), road and environment (\textit{Wthr\_Cond\_ID}, \textit{Light\_Cond\_ID}, \textit{Surf\_Cond\_ID}), crash dynamics (\textit{Intrsct\_Relat\_ID}, \textit{FHE\_Collsn\_ID}, \textit{Harm\_Evnt\_ID}), and temporal factors (\textit{Day\_of\_Week}, \textit{Crash\_Speed\_Limit}). Advanced safety features like \textit{HasAutomaticBrakingSystem} and \textit{HasAutomaticEmergencyBrakingSystem} are crucial for EV-specific severity prediction \cite{txdot_crash_records}.

\begin{figure*}[h!]
\centering
\includegraphics[width=1\linewidth]{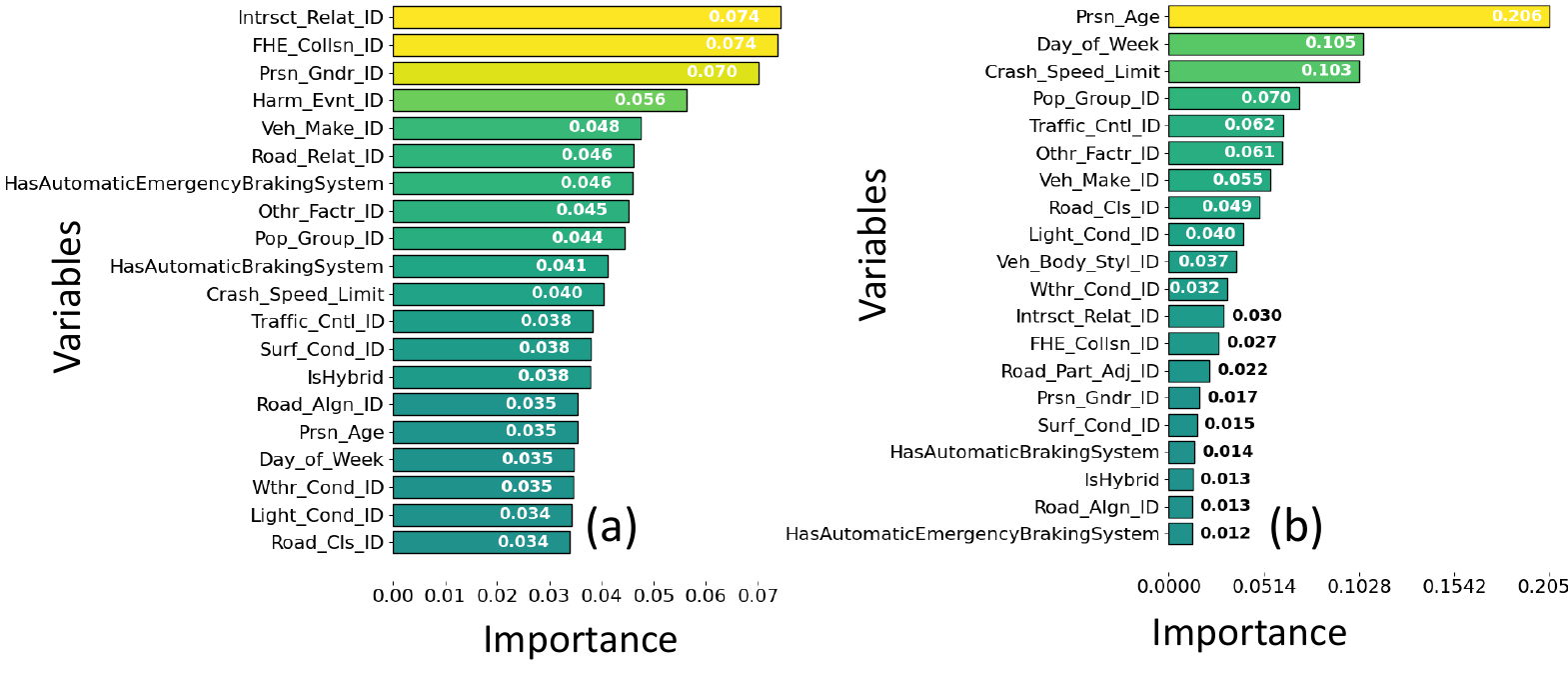}
\caption{Variable Selection Using XGBoost and Random Forest}
\label{fig:Variable_selection}
\end{figure*}

\subsection{Data Cleaning and Preprocessing}

Missing value treatment followed established protocols for crash data analysis, addressing the common issue of incomplete crash reporting. For categorical variables such as person demographics (\textit{Prsn\_Gndr\_ID, Prsn\_Rest\_ID}) and crash characteristics, mode imputation was applied to preserve the dominant category while retaining meaningful "Unknown" classifications. For numerical variables, including person age (\textit{Prsn\_Age}), vehicle model year (\textit{Veh\_Mod\_Year}), and speed limits (\textit{Crash\_Speed\_Limit}), median imputation was employed to minimize the influence of outliers or data entry errors.

To reduce sparsity and enhance model generalization, multiple categorical variables were consolidated into broader categories. For instance, person age was grouped into three major classes: $<30$, 31–60, and $>60$ years. Similarly, weather condition was reduced to six categories (clear, cloudy, rain, snow, fog, other), light condition into four (daylight, dark, dusk, other), and surface condition into five (dry, wet, gravel, snow, other). Crash speed limits and road alignment variables were also combined into logical bins to capture key distinctions while reducing noise. This grouping not only prevents overfitting caused by sparse category levels but also helps models learn robust patterns across semantically meaningful groups.

Categorical variable encoding used one-hot encoding for all nominal features, while ordinal features such as injury severity were encoded to preserve their ranking. Feature scaling was performed with StandardScaler for continuous variables (e.g., crash speed limits, person age), ensuring that variables of different magnitudes contribute equally during deep learning model training.

\subsection{Class Distribution and Resampling Strategy}

Crash severity data exhibited a significant class imbalance, with "No Injury" outcomes dominating over severe and fatal cases, which can bias models toward the majority class. To mitigate this, SMOTEENN, a hybrid technique combining SMOTE and Edited Nearest Neighbors, was applied. SMOTE generates synthetic samples for minority classes by interpolating between nearest neighbors\cite{husain2025smote}, while ENN removes noisy or overlapping majority class samples\cite{husain2025smote}. This two-step approach enhances minority class representation and has shown improved performance over SMOTE alone, particularly for neural networks and ensemble methods\cite{husain2025smote}.

\subsection{Feature Importance and Selection}

Feature importance analysis using XGBoost and Random Forest identified the top 10 common predictors of EV crash severity (Fig.~\ref{fig:Variable_selection}). XGBoost evaluates feature importance based on gradient-based gains, while Random Forest uses mean decrease in impurity \cite{scarano2025cyclist}. The combined ranking highlights key factors such as intersection relation, first harmful event, person age, crash speed limit, and day of week, alongside advanced safety features like automatic emergency braking systems. These results are consistent with prior research emphasizing the influence of driver demographics, crash context, and safety technologies on injury outcomes \cite{cicchino2017effectiveness, aukema2025study}.

\section{Methodology}
\subsection{Study Design}
The proposed framework for crash severity prediction consists of two primary stages: data preparation and modeling as seen in Fig. \ref{fig:flowchat}. In Stage~1, the raw dataset undergoes feature selection using Random Forest and XGBoost to identify the top 12 influential variables. A combined cleaning and resampling strategy using SMOTEENN and computed class weights is then applied to handle imbalanced classes, resulting in a resampled dataset.  

In Stage~2, severity prediction is performed using deep learning models, namely \textit{MambaNet}, \textit{MambaAttention}, and \textit{TabPFN} (pretrained with one-shot inference). The dataset is split into training (60\%), validation (20\%), and testing (20\%) subsets. Model performance is evaluated using accuracy, F1-score, precision, and recall metrics to ensure robust classification across severity levels (fatal, injury, no injury).

\begin{figure*}[h!]
\centering
\includegraphics[width=1.5\columnwidth]{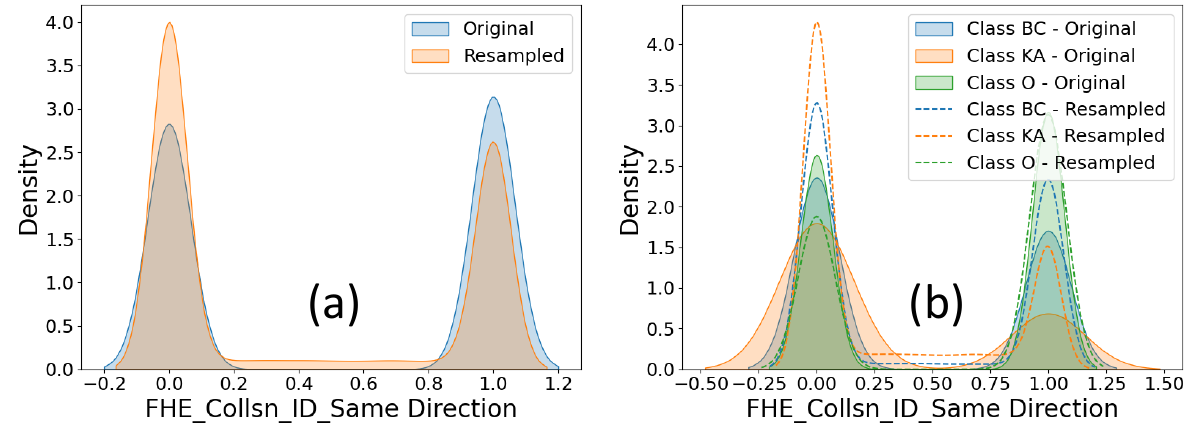}
\caption{Before and after feature distribution analysis with SMOTEENN (a)Feature distribution of first harmful event, same direction (b) Class-conditional feature distributions of first harmful event, same direction.}
\label{fig:SMOTEENN}
\end{figure*}

To address class imbalance in the crash severity dataset, the SMOTEENN technique was applied, combining synthetic oversampling with noise removal to create a more balanced distribution of KA, BC, and O crashes. Figure \ref{fig:SMOTEENN} illustrates the feature distribution of \textit{FHE\_Collsn\_ID\_Same Direction} before and after resampling. In Figure \ref{fig:SMOTEENN}a, the overall density remains consistent, while Figure \ref{fig:SMOTEENN}b shows improved class-conditional balance across severity levels. This demonstrates that SMOTEENN achieves class balancing without distorting key feature distributions.

\subsection{Hyperparameter Tuning}
Table~\ref{tab:model_config_short} summarizes the hyperparameter settings for the three models. The \textit{TabPFN} model does not require hyperparameter tuning, as it relies on a frozen transformer pretrained on 160M synthetic tasks, with only input and output dimensions specified for one-shot inference on GPU. The \textit{MambaNet} model uses two hidden layers ([128, 64]) with a dropout of 0.3, learning rate of 0.001, weight decay of 1e-4, and is trained for 50 epochs with batch size 64 using AdamW and a ReduceLROnPlateau scheduler. The \textit{MambaAttention} model adopts hidden layers [256, 128] with the same parameters but utilizes the Adam optimizer and a StepLR scheduler (step size = 10, gamma = 0.5).

\section{RESULTS AND DISCUSSIONS}
\subsection{Validation of Experiment}
MambaAttention’s 96\% accuracy stems from its attention mechanism and wider layers [256, 128], which better capture key crash features after SMOTEENN, improving class separation. MambaNet, with narrower layers and no attention, achieved 94\%. TabPFN, using one-shot inference with a frozen prior, could not fully adapt to the resampled feature distribution, resulting in 92\% (see Table~\ref{tab:training-validation}).

\begin{table}[H]
  \centering
  \scriptsize  
  \caption{Hyperparameter Tuning}
  \label{tab:model_config_short}
  \begin{tabular}{p{2cm} p{5cm}}
    \toprule
    \textbf{Model} & \textbf{Details} \\
    \midrule
    TabPFN & No traditional tuning. Uses a pretrained transformer on 160M synthetic tasks. \newline
    \textbf{Inputs:} \texttt{X\_train\_sel.shape[1]} $\rightarrow$ \texttt{len(label\_encoder.classes\_)}. \newline
    Inference is one-shot on \texttt{cuda} if available. \\
    \midrule
    MambaNet & \textbf{Dims:} [128, 64] $\rightarrow$ \texttt{len(classes)}. \newline
    \textbf{Dropout:} 0.3, \textbf{LR:} 1e-3, \textbf{WD:} 1e-4, \textbf{Epochs:} 50, \textbf{Batch:} 64. \newline
    \textbf{Opt:} AdamW, \textbf{Scheduler:} ReduceLROnPlateau. \\
    \midrule
    MambaAttention & \textbf{Dims:} [256, 128] $\rightarrow$ \texttt{len(classes)}. \newline
    \textbf{Dropout:} 0.3, \textbf{LR:} 1e-3, \textbf{WD:} 1e-4, \textbf{Epochs:} 50, \textbf{Batch:} 64. \newline
    \textbf{Opt:} Adam, \textbf{Scheduler:} StepLR (\texttt{step\_size=10, $\gamma$=0.5}). \\
    \bottomrule
  \end{tabular}
\end{table}

\begin{table}[ht]
\centering
\scriptsize  
\caption{Summary of Training and Validation for Crash Severity Prediction Models}
\label{tab:training-validation}
\begin{tabular}{l p{5cm}}
\toprule
\textbf{Model} & \textbf{Details} \\
\midrule
MambaAttention & Accuracy: 96\%. \newline
Epochs: 50 (Early Stopping). \newline
Number of Samples – KA: 19,907, BC: 19,046, O: 8,326. \\
\midrule
MambaNet & Accuracy: 94\%. \newline
Epochs: 50 (Early Stopping). \newline
Number of Samples – KA: 19,907, BC: 19,046, O: 8,326. \\
\midrule
TabPFN & Accuracy: 92\%. \newline
Epochs: N/A (One-Shot Inference). \newline
Number of Samples – KA: 19,907, BC: 19,046, O: 8,326. \\
\bottomrule
\end{tabular}
\end{table}

\subsection{Model Performance}
Table~\ref{tab:performance} compares the performance of \textit{MambaAttention}, \textit{MambaNet}, and \textit{TabPFN} across the crash severity categories KA, BC, and O using precision, recall, F1-score, and per-class accuracy. \textit{MambaAttention} demonstrates the strongest overall performance, achieving perfect precision and recall for KA and maintaining high accuracy for BC (95\%) and O (94\%). Its attention layers effectively capture the relative importance of key crash features such as collision type and roadway conditions, which enhances its capability to differentiate between severe and moderate injury classes, particularly after SMOTEENN balancing.

\textit{MambaNet}, leveraging its CNN-LSTM hybrid design, also excels in identifying KA crashes with perfect scores across all metrics (100\%) but performs slightly lower on O crashes (78\% accuracy), suggesting a challenge in distinguishing non-injury events with overlapping feature patterns. 

\textit{TabPFN} delivers competitive results for KA (97\% F1) and BC (90\% F1) but lags behind on O classification (84\% F1), primarily due to its reliance on one-shot inference with a frozen pretrained transformer that is not fine-tuned to the resampled EV crash dataset. These results highlight the benefits of architectures like \textit{MambaAttention} and \textit{MambaNet}, which dynamically learn feature interactions and sequential dependencies, achieving better generalization across severity categories.

\begin{table}[ht]
\centering
\caption{Prediction Performance of the Crash Severity Prediction Models}
\label{tab:performance}
\resizebox{\linewidth}{!}{
\begin{tabular}{lcccc}
\toprule
\textbf{Category} & \textbf{Precision (\%)} & \textbf{Recall (\%)} & \textbf{F1-Score (\%)} & \textbf{Accuracy (\%)} \\
\midrule
\multicolumn{5}{c}{\textbf{MambaAttention}} \\
KA & 100 & 100 & 98 & 98 \\
BC & 93  & 96  & 95 & 95 \\
O  & 90  & 84  & 87 & 94 \\
\midrule
\multicolumn{5}{c}{\textbf{MambaNet}} \\
KA & 100 & 100 & 100 & 99 \\
BC & 91  & 95  & 93  & 94 \\
O  & 87  & 79  & 82  & 78 \\
\midrule
\multicolumn{5}{c}{\textbf{TabPFN}} \\
KA & 97  & 97  & 97  & 94 \\
BC & 93  & 87  & 90  & 87 \\
O  & 79  & 90  & 84  & 90 \\
\bottomrule
\end{tabular}}
\end{table}

The \textit{MambaAttention} model outperformed the other two models in terms of consistency and robustness. It achieved perfect scores (100\%) for KA classification across all metrics, confirming its superior capacity to detect the most critical crash types. In the BC category, \textit{MambaAttention} reached a precision of 96\% and a recall of 95\%, with an F1-score of 96\% and class accuracy of 96\%. Although the performance dropped slightly for the O category, showing 88\% precision, 79\% recall, and 84\% accuracy, the model still maintained a high level of reliability.

Overall, \textit{MambaAttention} demonstrated the best balance of performance across all categories, particularly excelling in the severe and moderate injury classifications. \textit{MambaNet} also showed strong predictive capabilities, especially for severe cases, while \textit{TabPFN}, though efficient and lightweight, exhibited slightly reduced performance in distinguishing no-injury cases. These findings affirm the potential of Mamba-based architectures for crash severity prediction tasks, especially in scenarios where high accuracy in identifying life-threatening crashes is critical for real-time safety interventions.

 \begin{figure}[h!]
\centering
\includegraphics[width=1\columnwidth]{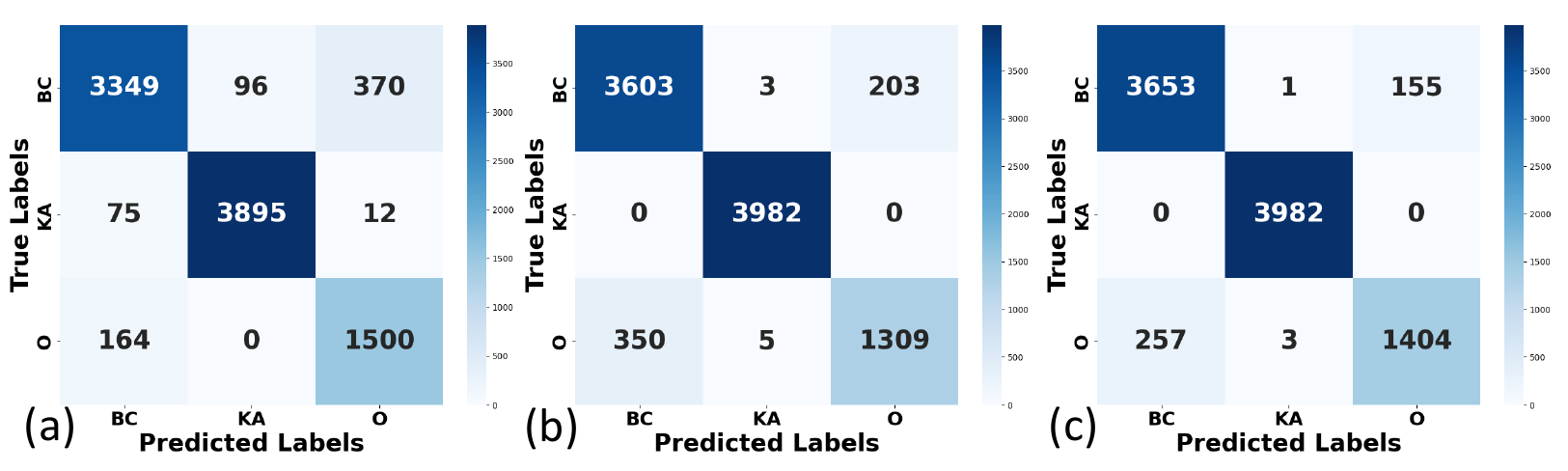}
\caption{Confusion Matrix of each model: (a) TabPFN; (b) MambaNet; (c) MambaAttention}
\label{fig:ConfusionMat}
\end{figure}

Figure~\ref{fig:ConfusionMat} illustrates the confusion matrices for \textit{TabPFN}, \textit{MambaNet}, and \textit{MambaAttention}, highlighting their classification behavior across KA, BC, and O categories. \textit{MambaAttention} (Fig.~\ref{fig:ConfusionMat}c) shows the most precise predictions, with minimal misclassifications, particularly for KA and BC classes, due to its attention mechanism that emphasizes the most informative crash features. \textit{MambaNet} (Fig.~\ref{fig:ConfusionMat}b) also achieves high accuracy but exhibits slightly higher confusion between O and BC cases, which is consistent with its lower recall for non-injury crashes in Table~\ref{tab:performance}. 

In contrast, \textit{TabPFN} (Fig.~\ref{fig:ConfusionMat}a) demonstrates moderate performance, with notable misclassifications between BC and O, reflecting its limited task-specific adaptation caused by one-shot inference. These confusion patterns align with the precision and F1-score trends in Table~\ref{tab:performance}, where \textit{TabPFN} underperforms on BC and O, while \textit{MambaAttention} maintains high scores across all classes. This comparison reinforces that attention-based and hybrid architectures reduce class overlap and improve overall classification reliability.

 \begin{figure}[h!]
\centering
\includegraphics[width=1\columnwidth]{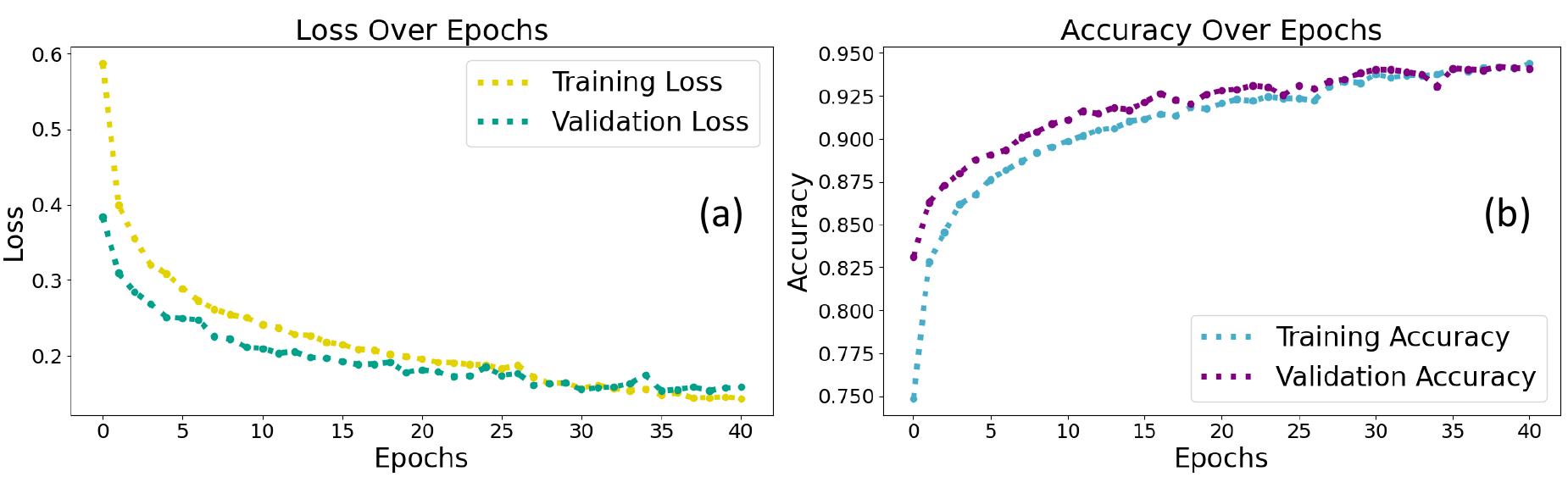}
\caption{Training performance of the MambaNet model: (a) Loss curves; (b) Accuracy curves}
\label{fig:MambaNet}
\end{figure}

\begin{figure}[h!]
\centering
\includegraphics[width=1\columnwidth]{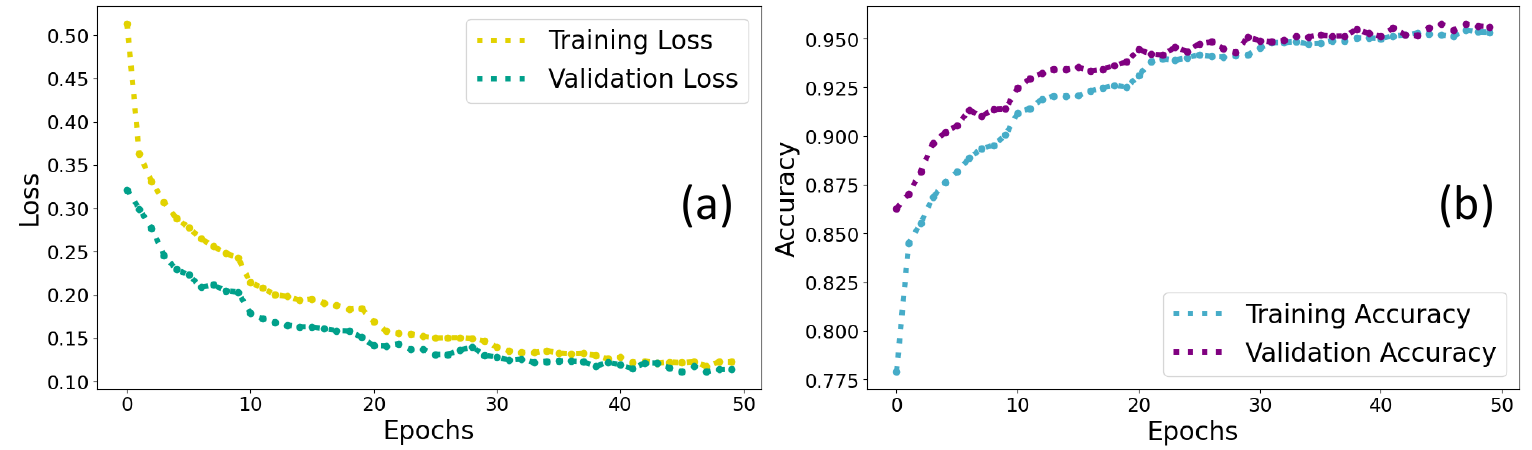}
\caption{Training performance of the MambaAttention model: (a) Loss curves; (b) Accuracy curves}
\label{fig:MambaAttention}
\end{figure}

Figures~\ref{fig:MambaNet} and \ref{fig:MambaAttention} illustrate the training behavior of the \textit{MambaNet} and \textit{MambaAttention} models, respectively. Both models show smooth convergence, with training and validation loss curves (Figs.~\ref{fig:MambaNet}a and \ref{fig:MambaAttention}a) steadily decreasing and stabilizing after approximately 30--35 epochs. The accuracy curves (Figs.~\ref{fig:MambaNet}b and \ref{fig:MambaAttention}b) demonstrate rapid initial improvements followed by consistent validation accuracy, indicating minimal overfitting due to the use of early stopping, dropout, and weight decay. 

\textit{MambaAttention} exhibits slightly lower final loss and faster convergence compared to \textit{MambaNet}, highlighting the advantage of its attention mechanism in capturing feature importance during training. In contrast, \textit{TabPFN} does not have learning curves since it operates in a one-shot inference mode using a frozen pretrained transformer, requiring no iterative training process. Overall, \textit{MambaAttention} delivered the most consistent and accurate predictions across all severity classes, supported by its attention mechanism that enhances feature importance modeling. \textit{MambaNet} also achieved strong results, especially for severe crashes, while \textit{TabPFN}, though efficient with one-shot inference, underperformed on BC and O classes due to limited dataset adaptation. The confusion matrices and learning curves confirm that Mamba-based models offer superior class separation and stable convergence, making them well-suited for crash severity prediction on structured EV datasets.

\section{Conclusions}
This study proposed a deep tabular learning framework for predicting crash severity in EV collisions using real-world data from Texas (2017–2023). We integrated SMOTEENN resampling with ensemble-based feature selection (XGBoost and Random Forest) to construct a compact, high-signal feature set for tabular classification. Among the evaluated models, \textit{MambaAttention} achieved the best performance (96\% accuracy), leveraging attention-based feature reweighting and sequential encoding to capture complex dependencies. \textit{MambaNet} followed closely, benefiting from its hybrid CNN-state space design, while \textit{TabPFN} offered efficient, zero-hyperparameter inference but exhibited limited adaptability to the resampled data. These results highlight the promise of state-space and attention-driven architectures in structured data domains, demonstrating their potential for interpretable and accurate prediction in safety-critical applications.

\bibliographystyle{IEEEtran}
\bibliography{main}

\end{document}